\documentclass{article}
\usepackage{spconf,amsfonts,amsmath,graphicx}
\usepackage{hyperref}
\hypersetup{
    colorlinks=true,
    linkcolor=magenta,
    filecolor=magenta,      
    urlcolor=magenta,
    pdftitle={Overleaf Example},
    pdfpagemode=FullScreen,
    }
\usepackage{booktabs,multirow}
\usepackage{xcolor}
\usepackage{colortbl}
\newlength\savewidth\newcommand\shline{\noalign{\global\savewidth\arrayrulewidth
  \global\arrayrulewidth 1.5pt}\hline\noalign{\global\arrayrulewidth\savewidth}}

\usepackage{enumitem}
\setlist{nosep, leftmargin=14pt}

\usepackage{mwe} 

\title{Dual Cross-Attention Siamese Transformer for Rectal Tumor Regrowth Assessment in Watch-and-Wait Endoscopy}
\name{
\begin{tabular}[c]{@{}l@{}}Jorge Tapias Gomez \textsuperscript{1} \quad Despoina Kanata \textsuperscript{2} \quad Aneesh Rangnekar\textsuperscript{1} \quad Christina Lee \textsuperscript{2}\\\quad \quad J. Joshua Smith \textsuperscript{2} \quad Julio Garcia-Aguilar \textsuperscript{2} \quad Harini Veeraraghavan \textsuperscript{1}\end{tabular}
}
\address{$^{1}$ Department of Medical Physics, Memorial Sloan Kettering Cancer Center, USA \\
$^{2}$ Department of Surgery, Colorectal Service, Memorial Sloan Kettering Cancer Center, USA}

\begin{document}
%
\maketitle
Increasing evidence supports watch-and-wait (WW) surveillance for patients with rectal cancer who show clinical complete response (cCR) at restaging following total neoadjuvant treatment (TNT). However, accurate methods to early detect local regrowth (LR) from follow-up endoscopy images during WW are essential to manage care and prevent distant metastases. Hence, we developed a Siamese Swin Transformer with Dual Cross-Attention (SSDCA) to combine longitudinal endoscopic images at restaging and follow-up and distinguish cCR from LR. SSDCA leverages pretrained Swin Transformers to extract domain agnostic features and enhance robustness to imaging variations. Dual cross attention is implemented to emphasize features from the paired scans without requiring any spatial alignment to predict response. SSDCA as well as Swin-based baselines were trained using image pairs from 135 patients and evaluated on a held-out set of image pairs from 62 patients. SSDCA produced the best balanced accuracy (81.76\% $\pm$ 0.04), sensitivity (90.07\% $\pm$ 0.08), and specificity (72.86\% $\pm$ 0.05). Robustness analysis showed stable performance irrespective of artifacts including blood, stool, telangiectasia, and poor image quality. UMAP clustering of extracted features showed maximal inter-cluster separation (1.45 $\pm$ 0.18) and minimal intra-cluster dispersion (1.07 $\pm$ 0.19) with SSDCA, confirming discriminative representation learning. Code and weights available at: \href{https://github.com/Jotanator/SSDCA}{https://github.com/Jotanator/SSDCA}

\section{Introduction}
\label{sec:introduction}

The Organ Preservation in Rectal Adenocarcinoma (OPRA) clinical trial showed that 47\% of patients with rectal adenocarcinoma treated with total neoadjuvant therapy (TNT) followed by a watch-and-wait (WW) surveillance avoid surgery and achieve sustained clinical response without reducing their chance of cure~\cite{OPRAH}. Subjective clinical assessment is only 65\% accurate, and suffers from large inter-rater variation~\cite{Felder2021EndoscopicFeature}, necessitating objectively accurate methods to distinguish sustained clinical response (cCR) from local regrowth (LR) for optimal patient management. 

\begin{figure}
    \centering
    \includegraphics[width=\columnwidth]{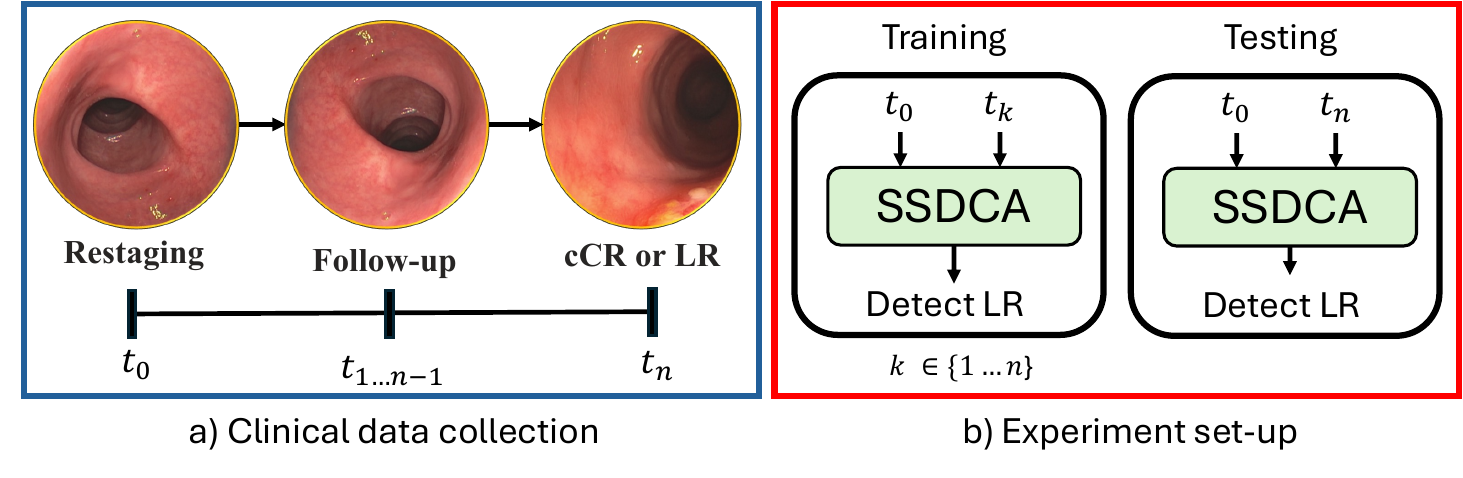}
    \caption{(a) Endoscopic images acquired every three months during watch-and-wait resulting in a longitudinal sequence from $t_0$ to $t_n$. (b) Experiment setup where paired images from two different timepoints are combined and provided as input into the Siamese Swin Dual Cross-Attention (SSDCA) model.}
    \label{fig:problemOverview}
\end{figure}

Although deep learning (DL) methods have shown capability to detect tumor response at restaging (8 to 12 weeks after TNT)~\cite{SPIE_Endo,Haak_Endo_ClinicVariables}, such methods are less accurate when applied to detect regrowing tumors that exhibit large variations in appearance, shape, and confounding factors such as blood~\cite{williams2024endoscopic}. However, an important feature in the management of patients on WW is the acquisition of multiple follow-up images until LR is detected or patients show sustained cCR for 3 years following TNT completion/restaging (Fig.~\ref{fig:problemOverview}a). Hence, we hypothesized that extracting the change in image features from restaging to follow-up would provide more accurate prediction of LR. 

Longitudinal endoscopic image analysis involves challenges including lack of spatial registration between images acquired at different times, variable number of follow ups due to differences in tumor regrowth patterns, and presence of varied artifacts such as blood, stool, telangiectasia, and scope that confound accurate detection~\cite{thompson2023}. Our contribution addresses these challenges by developing: (a) a Siamese architecture combining temporal features from restaging and follow-up images through bi-directional cross attention, obviating need for spatial alignment (Fig.~\ref{fig:problemOverview}b), (b) pretrained Swin encoders to enhance robustness to underlying image variations and we performed to study the impact of commonly occurring artifacts in endoscopic images, and (c) preliminary interpretability study to assess spatial similarity of the attentions from the two time points for classification. 

\section{Related works}

Multiple prior works have addressed the problem of predicting tumor treatment response by combining pairs of images either using radiomic~\cite{Sutton2020MachineLearningModel} or DL methods involving radiographic CT and MR images~\cite{Jin2021PredictingTreatmentResponse,Sun_LOMAI_2024,YUE2022102423}, demonstrating the benefit of combining additional time point in improving prediction accuracy. Methods to fuse information from temporal images vary, ranging from simple feature differencing~\cite{Sutton2020MachineLearningModel}, image or features concatenation~\cite{Jin2021PredictingTreatmentResponse} to cross-attention formulations~\cite{Sun_LOMAI_2024,YUE2022102423}. Long short term memory networks have also been used to model the temporal dynamics of feature changes between subsequent images acquired at fixed imaging intervals such as every week during radiation treatment~\cite{Xu2019DeepLearningPredicts}. However, aforementioned approaches require voxel-wise spatial correspondence between images achieved either through registration~\cite{Jin2021PredictingTreatmentResponse} or by focusing on region of interest centered on the detected tumors~\cite{Sun_LOMAI_2024}. Such spatial registration is harder to achieve for endoscopic images captured from potentially different view points and highly variable follow up times. Our approach addresses these gaps by using a dual cross attention which emphasizes relevant Swin Transformer features from both images to reliably detect LR.

\section{Methods}

\subsection{Datasets}

One hundred ninety seven patients with locally advanced rectal cancer, treated with induction or consolidation TNT and selected for WW surveillance at restaging, were analyzed in this study to classify LR versus cCR. A total of 2,278 images (LR $=$ 768, cCR $=$ 1,510) were available for use. We utilized 135 patients, corresponding to 7,392 image pairs (restaging and intermediate follow-up as well as temporally ordered pairs of follow-ups), for training our models. Testing was performed on a held-out set of 62 patients (368 image pairs of restaging and last follow up) as gold standard ground truth was only available after last follow up, when patients go for surgery or are determined to have sustained cCR. 

\subsection{Network architecture}

\begin{figure}[t]
\centerline{\includegraphics[width=0.48\textwidth]{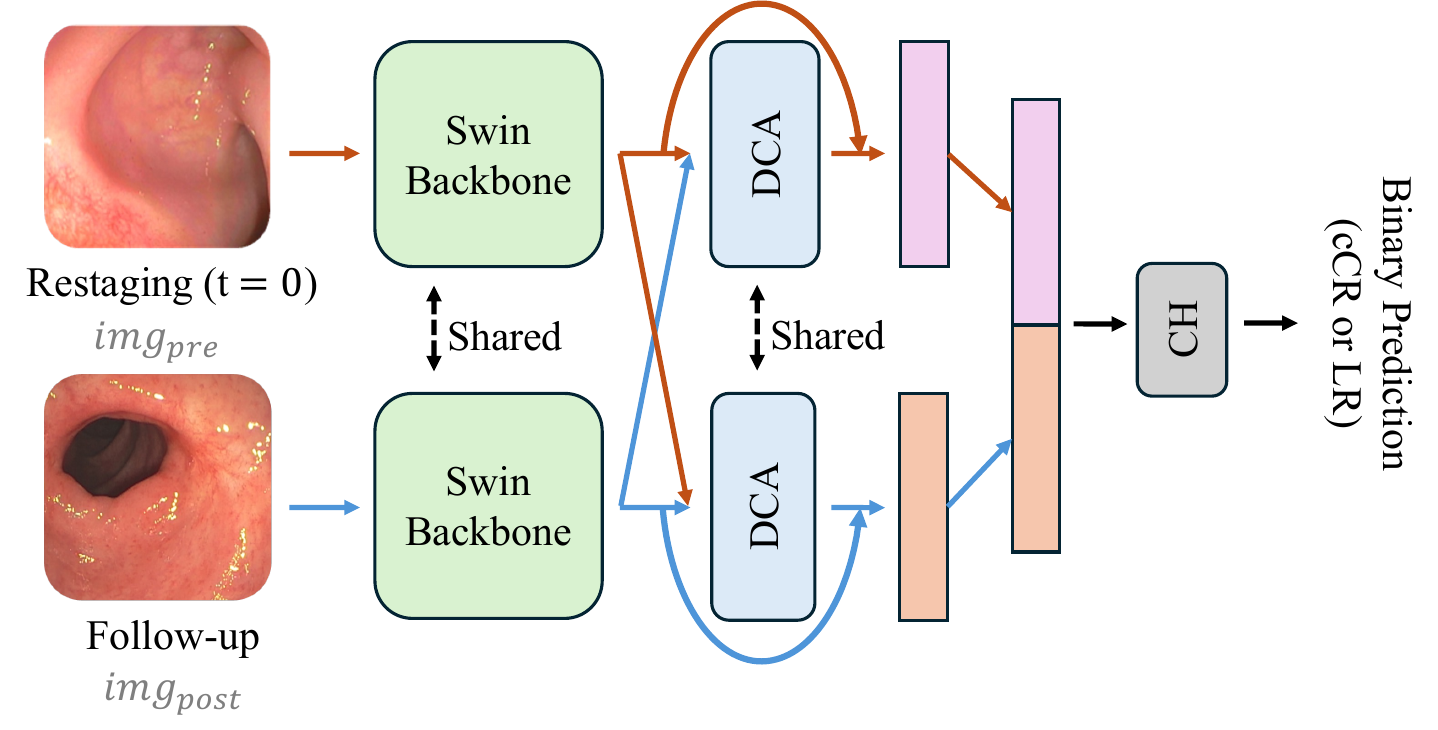}}
\caption{Architecture overview of the model.}
\label{Architecture}
\end{figure}

Fig.~\ref{Architecture} depicts an overview of our approach which combines Siamese neural networks with a Swin Transformer encoder to extract meaningful features from images taken at two different time points (refers to `SS'). A Swin encoder was used to alleviate the need to embed absolute positional information as is required in a vision transformer (VIT). Each time point first processes the endoscopic images independently, resulting in hierarchical features. To mitigate the impact of spatial misalignments and reduce the computational demands, we only focused on the features from Stage \#4 using the dual cross-attention (DCA) module. The DCA module emphasizes features across the time points, producing attention-refined residuals, which are then added back to the corresponding inputs. The resulting representation is then pooled, concatenated, and passed through a classification head (CH) to classify the pair as either LR or cCR. 

\subsection{Siamese Swin dual cross attention (SSDCA)}

A pair of longitudinal images, $x_{pre}$ and $x_{post}$, are processed by a shared Swin encoder $f_\theta$. The Stage \#4 features from the two images are extracted, where $F_{pre} = f^{(4)}_\theta(x_{pre})$ and $
F_{post} = f^{(4)}_\theta(x_{post})$, with shape $(H/32)\times(W/32)\times C$ (where $H$ and $W$ correspond to the spatial resolution, and $C$ to the number of channels). These are flattened and reshaped into matrices of size $(H/32 \cdot W/32)\times C$. The DCA module focuses on relevant features in one image by attending to semantically relevant features occurring in similar spatial locations in the other image using learned cross attention~\cite{CrossAttentionChangeDetection}. Specifically, dual cross attention is performed on $F_{pre}$ and $F_{post}$ by using features from one image as query, and the other as key and value pairs. Given $q_{pre} = F_{pre} W_q$, $q_{post} = F_{post} W_q$, $k_{pre} = F_{pre} W_k$, $k_{post} = F_{post} W_k$, $v_{pre} = F_{pre} W_v$, and $v_{post} = F_{post} W_v$, we obtained:

\begin{align}
\mathrm{CA}(F_{pre}) &= \mathrm{softmax}\!\left(q_{pre} k_{post}^\top / \sqrt{D_h}\right)v_{post},\\
\mathrm{CA}(F_{post}) &= \mathrm{softmax}\!\left(q_{post} k_{pre}^\top / \sqrt{D_h}\right)v_{pre}
\end{align}

\noindent,where $W_q,W_k,W_v\in\mathbb{R}^{C \times 3C_h}$ are the learned shared projection matrices. This ensured that the fused representation appropriately emphasizes spatially aligned features in both timepoints. Further, as such the relative spatial encoding focuses on extracting the local relationships and local spatial bias, it complements the cross attention extracted by the DCA to extract correspondences between the features in the two images. Residual connections and layer normalization (LN) are then applied:
\begin{align}
H_{pre} &= \mathrm{LN}\!\left(F_{pre} + \mathrm{CA}(F_{pre})\right)\\
H_{post} &= \mathrm{LN}\!\left(F_{post} + \mathrm{CA}(F_{post})\right)
\end{align}

\noindent Finally, global average pooling (GAP) is applied to $H_{pre}$ and $H_{post}$, and the
pooled features are concatenated and passed through the classification head:
\begin{align}
p = \sigma\!\Big(\mathrm{CH}\big(\operatorname{concat}(\mathrm{GAP}(H_{pre}),\,\mathrm{GAP}(H_{post}))\big)\Big)
\end{align}
where $\sigma$ denotes the sigmoid probability. This dual formulation yields a registration-free, correspondence-aware representation of the longitudinal changes between timepoints.

\subsection{Implementation details}

To ensure robust performance, we performed stratified 5-fold cross-validation on the training set (N $=$ 135 patients). For each fold, the model with the best validation accuracy was selected for evaluation. All five models were then evaluated on the test set (N $=$ 62 patients), and the reported results in this paper are the mean and standard deviation across all of them. 

All models were trained with cross-entropy loss with the Adam optimizer (learning rate: $2\times10^{-4}$) and a linear decay scheduler (warmup: N $=$ 10 epochs) with a batch size of 8 for 30 epochs. Random $90^\circ$ rotations, horizontal flips, and vertical flips were used as data augmentation. Balanced sampling was applied to mitigate class imbalance. All experiments were conducted in PyTorch~1.13.1 on a single A40 NVIDIA GPU.

We used the Swin-Small variant with a patch size of $4\times4$ and window size of $7\times7$ for $224\times224$ inputs, initialized with ImageNet-pretrained weights. In addition, two Swin-Small based variants were benchmarked: (1) consisting of just the single image (Swin-S~SI), which served as the non-Siamese baseline~\cite{SPIE_Endo}, and (2) a simple feature concatenation-based approach (SSFC), which performed feature concatenation at Stage \#4. Both variants were trained and evaluated with identical data as SSDCA.

\begin{table}[t]
\centering
\caption{Testing data accuracy for analyzed models (averaged across 5-fold cross-validation).}
\def\arraystretch{1.25}
\label{tab:tsne}
\resizebox{0.48\textwidth}{!}{%
\begin{tabular}{lccc}
Model & Balanced Accuracy & Sensitivity & Specificity \\
\shline
Swin-S SI & 76.24 $\pm$ 0.02 & 65.32 $\pm$  0.09 & 87.14 $\pm$ 0.07\\
SSFC & 81.13 $\pm$ 0.06 & 84.00 $\pm$ 0.13& 78.57 $\pm$ 0.12\\
SSDCA & 81.76 $\pm$  0.05 & 90.07 $\pm$ 0.08 & 72.86 $\pm$ 0.05\\
\bottomrule
\end{tabular}%
}
\end{table}

\subsection{Metrics}

We reported balanced accuracy, specificity, and sensitivity in this paper. For patients with multiple images acquired on the same day (restaging and follow-up), predictions were aggregated using a top-$k$ strategy (k $=$ 3), averaging the probabilities from the three most confident combinations. Robustness to common imaging artifacts such as telangiectasia (TLG), blood, stool, and poor quality (PQ) that includes specular reflections, image over or under-saturation, occlusions from presence of scope was analyzed by computing specificity and sensitivity under these conditions.

\section{Results}

\subsection{SSDCA more accurately predicts LR}
As shown in Table ~\ref{tab:tsne}, SSDCA was the most accurate among all methods. SSDCA had higher sensitivity than SSFC but lower specificity than the latter method. The single image prediction method (Swin-S SI) was the least accurate, showing lower balanced accuracy than SSDCA and SSFC.

\subsection{SSDCA features are better separated indicating better discrimination for classification}

\begin{figure}[t]
\centerline{\includegraphics[width=\linewidth]{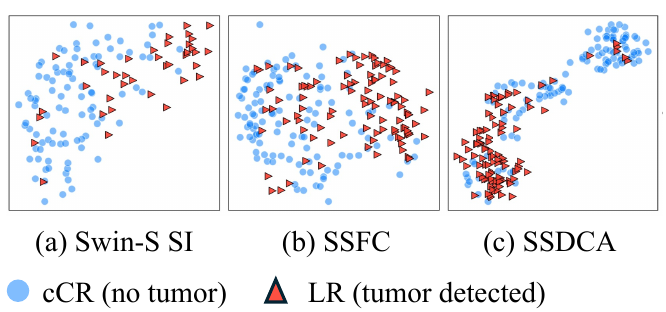}}
\caption{UMAP visualization of feature embeddings from the testing set. Each point represents an image combination (or a single image for Swin-S SI), color-coded by final clinical outcome. LR = local regrowth, cCR = complete clinical response.}
\label{PCA}
\end{figure}

\begin{figure}[t]
\centerline{\includegraphics[width=\linewidth]{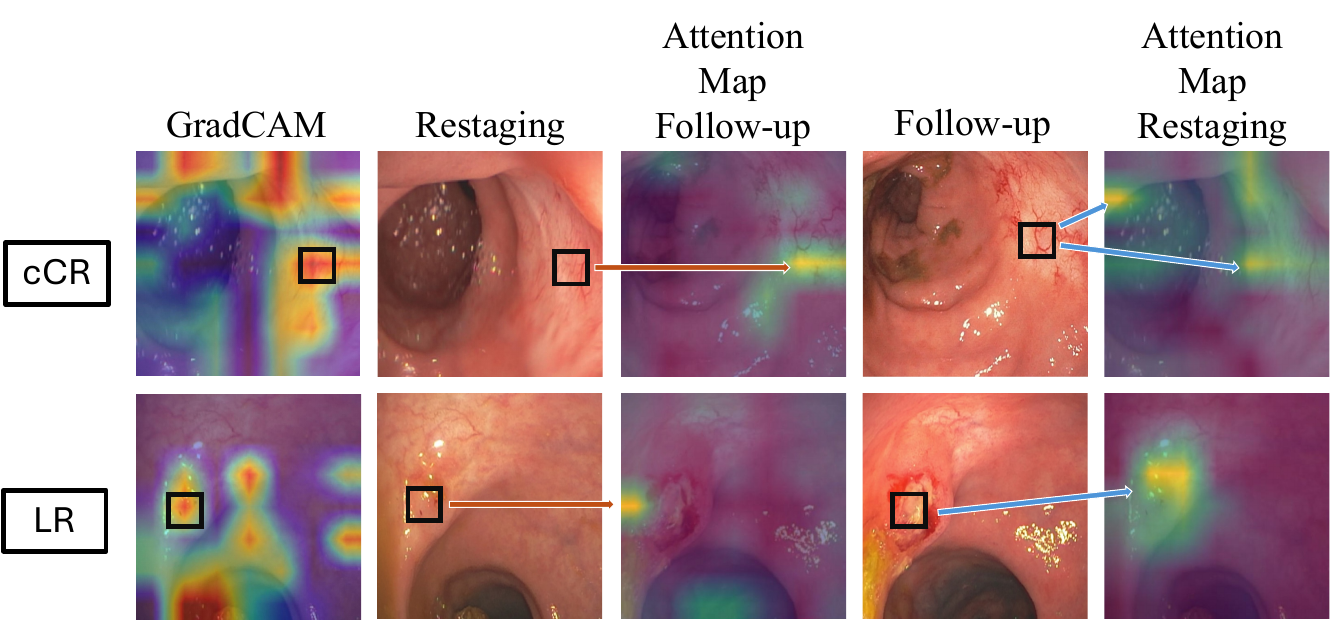}}
\caption{GradCAM and attention maps for two representative test cases produced using SSDCA shows good correspondence of relevant spatial features between the images.}
\label{Attention}
\end{figure}

We employed Uniform Manifold Approximation and Projections (UMAPs) for visualizing the features and computed inter/intra cluster distances in the original feature space using PCA for this analysis. Fig.~\ref{PCA} shows the rendering of features (post DCA module) computed from the test cases using the different models. As observed, SSDCA produced the best separation of patients by outcomes, achieving the highest inter-cluster distance (1.45 $\pm$ 0.18) and lowest intra-cluster spread (1.07 $\pm$ 0.19). In contrast, both Swin-S~SI and SSFC exhibited reduced separation (1.18 $\pm$ 0.21 and 1.38 $\pm$ 0.19, respectively) and large spatial spread of the clusters  (1.10 $\pm$ 0.06 and 1.22 $\pm$ 0.09). On the other hand, SSDCA resulted in tighter margins (\,$\approx$\,{+}0.25 inter,\,--0.15 intra) confirming why better discrimination of patient outcomes resulted from this method.

\subsection{Attention map correspondences between temporal image pairs with SSDCA}

Fig.~\ref{Attention} visualizes the spatial correspondences between the attended features in the restaging and the follow-up images for two representative test cases. We used the Gradient-weighted Class Activation Mapping (GradCAM) to extract the relevant regions used for prediction using the restaging image. A sample selected region with high attention shows correspondingly high activation in the attention map from the restaging image, despite lack of any spatial alignment between the two images. Finally, the high attention region from follow-up image was used to compared to the attended region in the restaging, which shows strong correspondence for the LR case (second row) and high activation in the corresponding as well as additional regions for the cCR case. The selected cases contained variations also due to presence of blood and specular reflection from mucosa.

\subsection{Robustness analysis across artifacts}

\begin{figure}[t]
\centerline{\includegraphics[width=\linewidth]{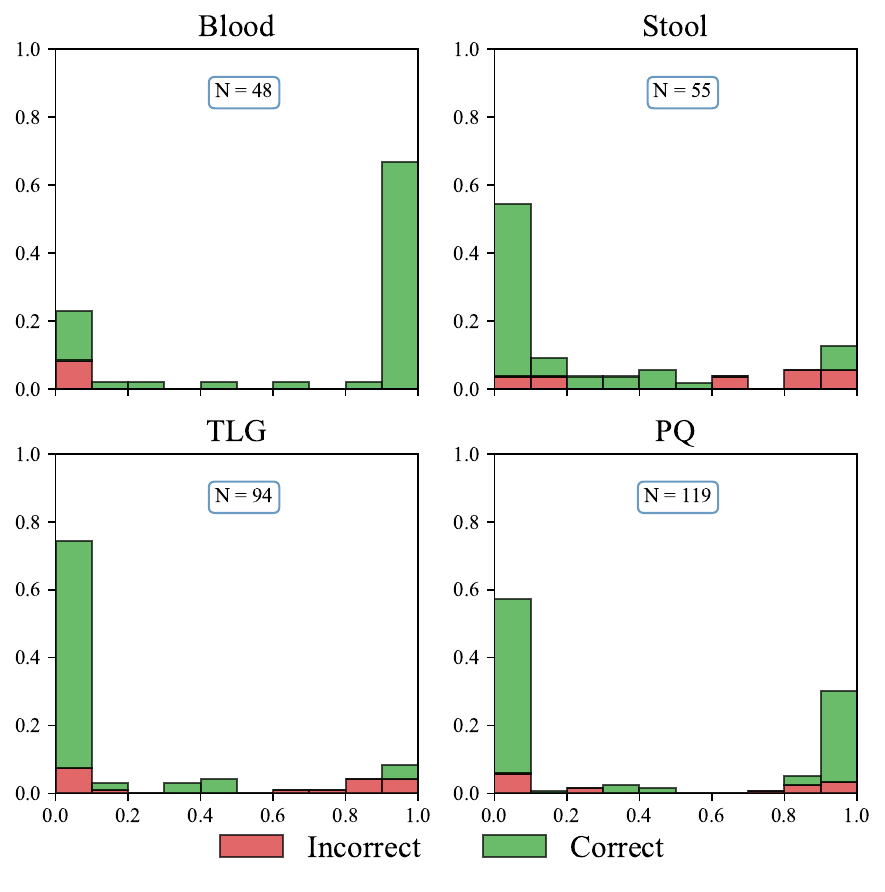}}
\caption{SSDCA output probability distributions under different imaging variations (blood, stool, TLG, and PQ). Green represents the correctly predicted sample, and red the incorrectly predicted samples.}
\label{Robustness}
\end{figure}

Fig.~\ref{Robustness} shows the output probability distributions of the SSDCA model under analyzed imaging artifacts. As shown, SSDCA was robust to presence of blood in identifying tumors, but resulted in lower probability for detecting tumors in the presence of stool, which occasionally obscured tumors and TLG, a known marker of cCR~\cite{thompson2023}. Under PQ, the model tended to result in incorrect detections for both LR and cCR.

\subsection{Impact of stage used for DCA}

Impact of computing DCA with the different stages of the Swin backbone (stages 1-4) was evaluated. Fusion at stage \#4 and \#2 achieved the highest balanced accuracy (82\% $\pm$ 0.05 and 82\% $\pm$ 0.01), followed very closely by stage \#3 (81\% $\pm$ 0.07), while stage \#1 produced (74\% $\pm$ 0.07). These results indicate that deeper stages produced higher accuracy possibly due to better discrimination ability.



\section{Discussion and conclusion}


We developed a longitudinal analysis framework to predict tumor regrowth from endoscopic images of rectal cancer patients. Our findings show that combining pairs of images with dual cross attention leads to more accurate performance compared to single timepoint analysis. SSDCA balanced accuracy of 82\% is comparable to surgeons' accuracy ranging between 74\% to 83\% to detect LR~\cite{williams2024endoscopic}. Our analysis also showed that DCA extracted spatially similar features despite lack of spatial alignment between images to inform prediction. Finally, SSDCA was reasonably robust to common endoscopic artifacts, suggesting that it is a reasonable approach for endoscopic image based prediction of treatment response.  

\section{Compliance with Ethical Standards}
This retrospective research study was conducted in line with the principles of the Declaration of Helsinki. Approval was granted by the Ethics Committee of Memorial Sloan Kettering Cancer Center.

\section{Acknowledgments}

This research was supported by Department of Surgery at Memorial Sloan Kettering. We thank Maria Widmar, Iris H Wei, Emmanouil P Pappou, Garrett M Nash, Martin R Weiser, and Philip B Paty, along with Hannah Thompson, Hannah Williams, Joshua Jesse Smith and Julio Garcia-Aguilar, for their assistance in collecting endoscopic images.

\bibliographystyle{IEEEbib}
\bibliography{references}

\end{document}